\documentclass[letterpaper]{article} 
\usepackage{aaai2027}  
\nocopyright
\usepackage[hyphens]{url}  
\usepackage{graphicx} 
\def\UrlFont{\rm}  
\usepackage{natbib}  
\usepackage{caption} 
\usepackage{amsmath,amssymb}
\usepackage{algorithm}
\usepackage{algorithmic}
\usepackage{tabularx}
\usepackage{multirow}
\usepackage{xcolor}
\usepackage{colortbl}
\usepackage{bm}
\newcommand{\cmark}{\checkmark}
\usepackage{makecell}
\definecolor{VenueText}{RGB}{88,88,88}
\newcommand{\venue}[1]{\,{\scriptsize\textcolor{VenueText}{[#1]}}}

\usepackage{newfloat}
\usepackage{listings}
\lstdefinestyle{promptbox}{%
    basicstyle=\footnotesize\ttfamily,
    numbers=none,
    backgroundcolor=\color{black!6},
    frame=single,
    framerule=0pt,
    framesep=6pt,
    framexleftmargin=4pt,
    framexrightmargin=4pt,
    framextopmargin=4pt,
    framexbottommargin=4pt,
    xleftmargin=0pt,
    xrightmargin=0pt,
    columns=fullflexible,
    keepspaces=true,
    showstringspaces=false,
    breaklines=true,
    breakatwhitespace=false,
    tabsize=2,
    aboveskip=4pt,
    belowskip=6pt
}
\DeclareCaptionStyle{ruled}{labelfont=normalfont,labelsep=colon,strut=off} 
\floatstyle{ruled}
\newfloat{listing}{tb}{lst}{}
\floatname{listing}{Listing}

\usepackage{booktabs}
\colorlet{ResultHighlight}{blue!10}
\colorlet{HeaderAES}{red!15}
\colorlet{HeaderCM}{orange!25}
\colorlet{HeaderSC}{green!20}
\colorlet{HeaderRG}{cyan!20}
\title{StyleForge: Indoor Furniture Styling by Counterfactual Reasoning in a Hypergraph Field}

\author{
    Lingwei Dang\textsuperscript{\rm 1}\equalcontrib, 
    Shishuo Shang\textsuperscript{\rm 1}\equalcontrib, 
    Pan Liu\textsuperscript{\rm 1}\equalcontrib, 
    Jiajia Cheng\textsuperscript{\rm 1}, 
    Ziyan Qiu\textsuperscript{\rm 1}, 
    Zhenhao Zhang\textsuperscript{\rm 2}, \\
    Yufei Zhu\textsuperscript{\rm 2}, 
    Shenghui Huang\textsuperscript{\rm 1}, 
    Qingxin Xiao\textsuperscript{\rm 1}, 
    Yun Hao\textsuperscript{\rm 1}, 
    Juntong Li\textsuperscript{\rm 1}, 
    Qingyao Wu\textsuperscript{\rm 1}\thanks{Corresponding author.}
}
\affiliations{
    \textsuperscript{\rm 1}School of Software Engineering, South China University of Technology\\
    \textsuperscript{\rm 2}School of Information Science and Technology, ShanghaiTech University \\
    \{levondang, liupan\_xy\}@163.com, shangshishuo588@gmail.com, qyw@scut.edu.cn
}

\begin{document}
\maketitle

\begin{abstract}
Fixed-layout indoor furniture styling requires selecting assets that form a coherent room without changing the prescribed furniture categories, positions, orientations, or scales. Existing approaches typically retrieve each asset independently or rely on static local relations, making them prone to shape, material, and color conflicts after scene composition. We introduce StyleForge, a scene-level structured selection framework built on a dynamic hypergraph style field. A frozen multimodal large language model extracts structured style priors from an open-ended style request and the fixed layout, while StyleForge maintains a learnable candidate distribution for each furniture slot. Conditioned on the target style, the dynamic hypergraph style field adaptively activates and weights layout-induced hyperedges to capture higher-order dependencies among furniture. Counterfactual style preference learning then treats each candidate as a local substitution in the current style field and evaluates its contextual compatibility using Mahalanobis energies. Training alternates between optimizing the style field and the candidate logits. At inference, the model remains frozen and test-time training updates only room-specific candidate logits, progressively correcting cross-slot style conflicts as the global scene context evolves. Experiments on 3D-FRONT demonstrate state-of-the-art furniture retrieval and scene-level style coherence, producing more coherent fixed-layout furniture arrangements than object- and scene-level retrieval baselines.
\end{abstract}


\begin{figure}[t]
    \centering
    \includegraphics[width=0.95\columnwidth]{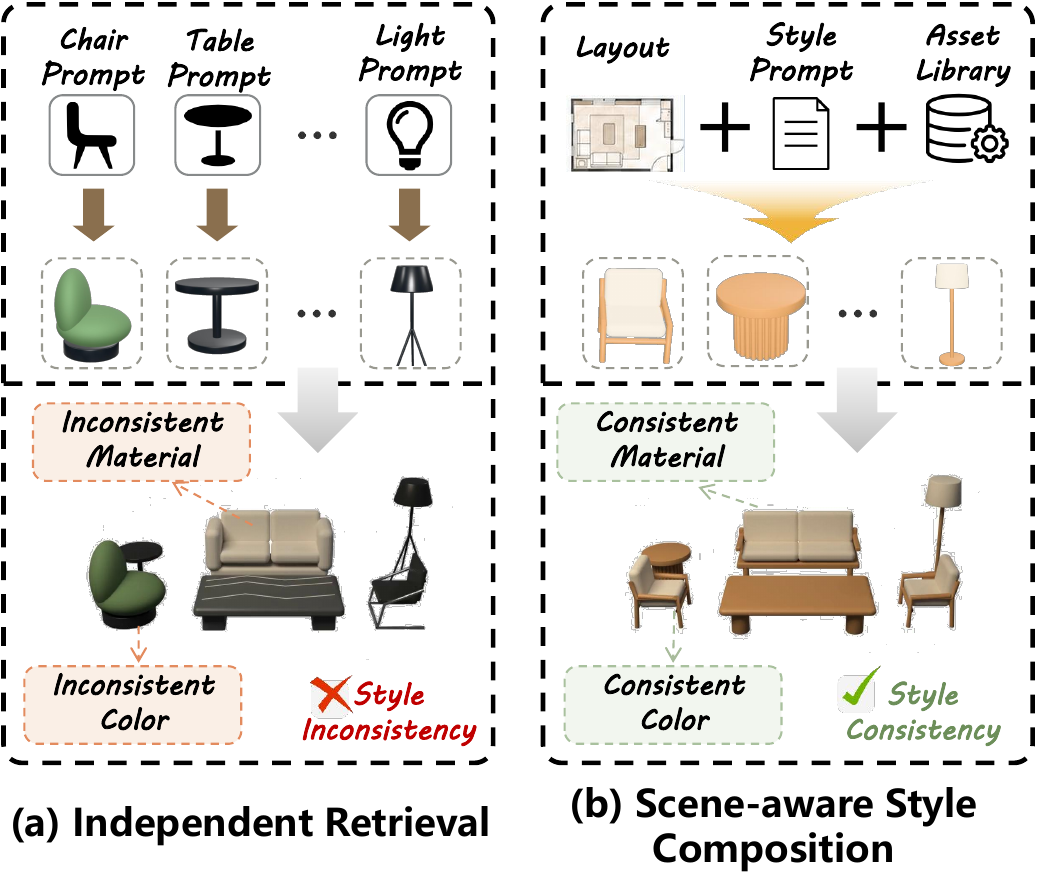}
    \caption{Comparison of two fixed-layout furniture-styling paradigms.
    (a) Independent object-level matching can combine individually relevant
    assets into a scene with shape, material, and color conflicts. (b) StyleForge
    jointly optimizes all slots to produce a coherent room-level assignment.}
    \label{fig:top}
\end{figure}

\section{Introduction}
\label{sec:introduction}

An asset that matches a target style in isolation may become conspicuously incompatible when placed with other furniture in a three-dimensional room. This observation motivates fixed-layout indoor furniture styling: given a target-style description, a prescribed room layout, and a large asset library, the system must assign one asset to each furniture slot while preserving its category, position, orientation, and scale. Unlike general indoor scene synthesis, this task cannot avoid an incompatible combination by changing the spatial arrangement. Its fundamental challenge is therefore to bridge object-level semantic relevance and scene-level aesthetic coherence. Addressing this challenge is important for virtual interior design, 3D content creation, and immersive embodied environments.

Existing approaches mainly follow two paradigms. Object-level cross-modal retrieval methods use pretrained representations such as CLIP, ULIP, and OpenShape \citep{radford2021learning,xue2023ulip,liu2023openshape} to measure the similarity between a style description and an individual asset. Although these representations retrieve semantically relevant furniture, they cannot determine whether an asset remains appropriate after scene composition, where quality also depends on its color, material, form, and spatial relation to surrounding furniture. Relation-based methods instead model a scene with graphs or hypergraphs \citep{wald2020learning,lv2024sgformer,feng2019hypergraph,jiang2019dynamic}. However, most rely on fixed pairwise edges or a static topology and therefore cannot express higher-order furniture dependencies that vary with the layout and requested style. Both paradigms also typically terminate after a single retrieval or ranking pass, without revising local choices according to the evolving scene context.

Our key insight is that an isolated furniture asset carries only partial style semantics, whereas room style emerges from the spatial, functional, and global relations among multiple assets. Fixed-layout furniture styling should therefore be formulated as scene-level structured selection rather than a collection of independent retrieval problems. Under this formulation, the preferred candidate for one slot depends on the current choices at other slots, while each local change alters the global style context.

Based on this insight, we propose StyleForge, a counterfactual reasoning framework built on a dynamic hypergraph style field. A frozen multimodal large language model extracts structured room- and slot-level style priors from the open-ended request and fixed layout. Instead of committing each slot to a top-ranked asset, StyleForge maintains a distribution over its candidates throughout optimization, allowing scene context to continuously reshape local preferences. The dynamic hypergraph style field represents furniture slots as distributional nodes and conditionally activates and weights layout-induced hyperedges according to the target style. Local hyperedges capture spatially or functionally related furniture groups, while a global hyperedge summarizes the room, enabling higher-order style dependencies to emerge from local-to-global propagation.

To convert scene compatibility into candidate preferences, we introduce counterfactual style preference learning. Each candidate is treated as a local substitution in the current style field while the remaining slot distributions are held fixed. Candidate- and scene-level Mahalanobis energies then measure the contextual compatibility of the resulting configuration. Their learned dimension-wise tolerances distinguish restrictive style cues from weakly relevant variation, providing an adaptive measure of conflicts in color, material, and form. StyleForge alternates between two optimization stages during training: with candidate distributions fixed, preference ranking learns an energy field that assigns lower energy to coherent configurations; with the field fixed, the resulting counterfactual energies update the candidate logits. At inference, the multimodal large language model and learned style field remain frozen, and test-time training updates only the room-specific candidate logits. The furniture assignment can consequently improve with its evolving global context instead of terminating after one-shot retrieval.

Experiments on 3D-FRONT show that StyleForge achieves state-of-the-art furniture retrieval and scene-level style coherence. Ablation and qualitative analyses further support the effectiveness of dynamic higher-order reasoning, Mahalanobis energy, and iterative refinement.

Our contributions are as follows:

\begin{enumerate}
\item We introduce a \textbf{dynamic hypergraph style field} that represents furniture slots as candidate distributions and uses style-conditioned hyperedge activation and propagation to capture local-to-global higher-order style dependencies.
\item We propose \textbf{counterfactual style preference learning}, which evaluates each candidate as a local substitution in the current style field and uses candidate- and scene-level Mahalanobis energies to measure its contextual compatibility.
\item We develop an \textbf{alternating optimization and test-time training strategy} that learns the style field and candidate logits in alternating stages, then freezes the field and iteratively refines room-specific candidate distributions at inference.
\end{enumerate}

\section{Related Work}
\label{sec:related_work}
\paragraph{3D Asset Retrieval.}
Vision--language pretraining provides shared semantic spaces for open-vocabulary 3D asset retrieval. CLIP establishes transferable image--text representations, while ULIP, ULIP-2, OpenShape, and Uni3D align language, images, and point clouds for scalable 3D representation learning \citep{radford2021learning,xue2023ulip,xue2024ulip,liu2023openshape,zhou2024uni3d}; Point-Bind and OpenScene extend multimodal alignment to richer 3D understanding \citep{guo2023point,peng2023openscene}. These methods primarily score text--asset relevance and cannot determine whether locally plausible furniture remains coherent after composition in color, material, and form. Prior work learns style from furniture compatibility, real-world co-occurrence, or scene images \citep{liu2015style,liu2019learning,weiss2020image}, and exploits local context or joint layout--object encodings for recommendation \citep{savva2017scenesuggest,pan2025metafind}. StyleForge instead represents all slots as coupled candidate distributions and jointly optimizes the complete assignment under a shared higher-order style field.

\begin{figure*}[t]
    \centering
    \includegraphics[width=\textwidth]{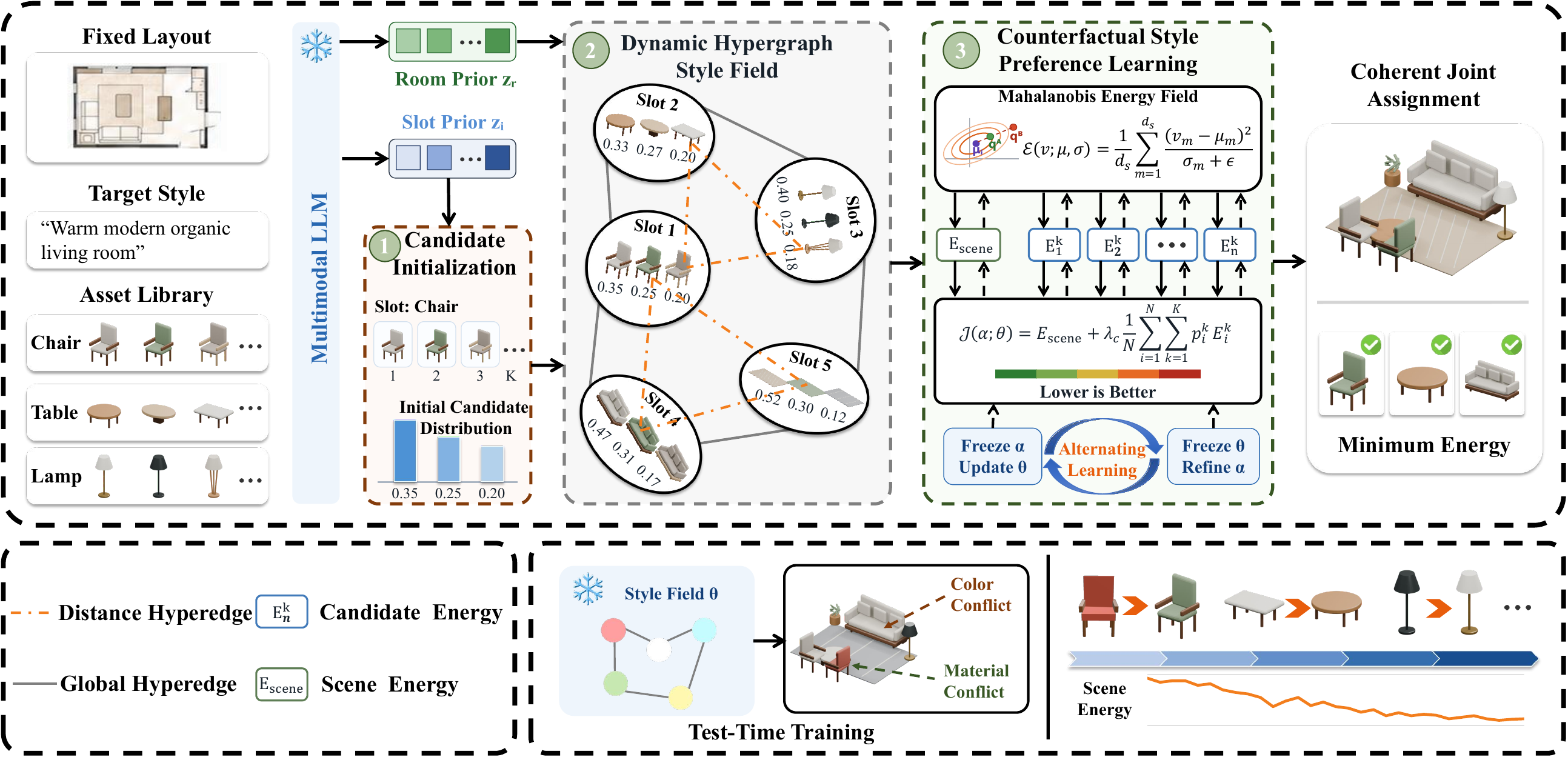}
    \caption{Overview of StyleForge. A frozen MLLM generates structured style
    priors and initializes candidate distributions. A dynamic hypergraph
    propagates higher-order context, while counterfactual substitutions are
    scored by Mahalanobis energy. Training alternates field and logit updates;
    TTT optimizes room-specific candidate logits.}
    \label{fig:framework}
\end{figure*}

\paragraph{Indoor Scene Synthesis.}
Indoor scene synthesis has evolved from example-based arrangements and relation-graph modeling \citep{fisher2012example,wang2019planit} to autoregressive generation on large furnished-scene datasets \citep{fu20213d,paschalidou2021atiss}. Recent approaches use scene graphs, diffusion, and structured generation or editing to model layout, geometry, and appearance \citep{zhai2023commonscenes,tang2024diffuscene,ju2024diffindscene,zhao2024roomdesigner,lin2024instructscene,zhai2024echoscene}. Large language and vision--language models further support instruction following, commonsense constraints, and style-conditioned planning \citep{yang2024holodeck,feng2023layoutgpt,sun2025layoutvlm,marshall2025decorum,pan2026metaspatial,berdoz2026text}, while related work studies multimodal generation, implicit representations, editable synthesis, and texturing \citep{yang2025mmgdreamer,liang2025s,zheng2025editroom,hollein2023text2room,huang2025roompainter}. These methods generate or modify object categories, poses, geometry, or appearance. We address a complementary, more constrained problem: selecting real library assets that jointly realize a target style while categories, positions, orientations, and scales remain fixed.

\paragraph{Higher-Order Structured Inference.}
Scene graphs encode object relations for 3D scene understanding \citep{wald2020learning,lv2024sgformer}, but fixed pairwise edges cannot fully express the groupwise material, color, form, and spatial relations that define indoor style. Hypergraph neural networks model multi-node relations, and dynamic variants adapt higher-order connectivity to the input \citep{feng2019hypergraph,jiang2019dynamic}. Recent work further studies dynamic and multi-hop hypergraph reasoning \citep{zhou2023totally,xie2025k,li2025dvhgnn}; in other structured prediction domains, multi-scale graph inference and distributional candidate generation model cross-node dependencies and output uncertainty \citep{dang2021msr,dang2022diverse}. In parallel, energy-based learning assigns low energy to compatible configurations \citep{lecun2006tutorial}. Structured prediction energy networks and iterative energy minimization support global inference over coupled outputs \citep{belanger2016structured,du2022learning}, while compositional energy models combine multiple constraints \citep{du2020compositional}. StyleForge applies these ideas to style-aware asset selection: the target style activates and weights layout-induced hyperedges, nodes represent candidate distributions rather than fixed features, and learned Mahalanobis energies evaluate counterfactual local substitutions.

\section{Method}
\label{sec:method}

\subsection{Problem Formulation and Overview}

Given a fixed indoor layout, a target-style description, and an asset library $\mathcal{A}$, our goal is to assign one asset to each of $N$ furniture slots while preserving the prescribed category, position, orientation, and scale of every slot. Slot $i$ is specified by its category $c_i$ and geometric descriptor $g_i$. After category filtering, we retrieve a candidate set $\mathcal{C}_i=\{x_i^k\}_{k=1}^{K}$. The desired output is a joint assignment $\hat{\mathcal{X}}=\{x_i^{\hat{k}_i}\}_{i=1}^{N}$ that agrees with the style request both individually and as a scene.

StyleForge instead maintains a categorical candidate distribution $p_i=\operatorname{softmax}(\alpha_i)$ for every slot and optimizes all distributions jointly. As shown in Fig.~\ref{fig:framework}, a frozen MLLM extracts structured room- and slot-level style priors from the request and layout. A dynamic hypergraph style field then propagates higher-order context, while counterfactual Mahalanobis energies evaluate each candidate as a local substitution. Training alternates between learning the style field and refining candidate logits; at test time, all model parameters are frozen and only the room-specific candidate logits are updated.

\subsection{Style-Prior Candidate Initialization}

\textbf{Structured style prior.}
We prompt the frozen MLLM with the target
description and fixed layout to produce room-level and slot-specific
retrieval descriptions. Their embeddings $z_r$ and $z_i$ encode the
shared palette, material, and form language and its category-specific
realization, respectively.

\textbf{Category-aware retrieval.}
A frozen multimodal encoder indexes multi-view asset renders.
For slot $i$, category-constrained retrieval returns the top-$K$
candidates and similarities $s_i^k$, from which we initialize

\begin{equation}
\left\{
\begin{aligned}
p_i^{k,0}
&=\frac{\exp(s_i^k/\tau)}
{\sum_{j=1}^{K}\exp(s_i^j/\tau)},\\
\alpha_i^{k,0}
&=\gamma\log\!\left(p_i^{k,0}+\epsilon\right),\\
p_i&=\operatorname{softmax}(\alpha_i),
\end{aligned}
\right.
\label{eq:candidate_initialization}
\end{equation}
where $\tau$ is the retrieval temperature, $\gamma$ controls initialization sharpness, and $\epsilon$ ensures numerical stability. Retaining a distribution rather than a single retrieval allows scene-level reasoning to revise locally plausible but globally incompatible choices.

\subsection{Dynamic Hypergraph Style Field}

\textbf{Distributional slot nodes.} Each node represents a furniture slot . For the candidate asset $x_i^k, f_i^k \in \mathbb{R}^{d_f}$ denotes its fixed visual-semantic feature. Given $p_i$, its soft asset representation, uncertainty, and initial state are

\begin{equation}
\left\{
\begin{aligned}
\bar f_i&=\sum_{k=1}^{K}p_i^k f_i^k,\\
H(p_i)&=-\sum_{k=1}^{K}p_i^k\log p_i^k,\\
h_i^0&=\phi_n\!\left([\bar f_i,g_i,z_i,z_r,H(p_i)]\right),
\end{aligned}
\right.
\label{eq:slot_representation}
\end{equation}
where $\phi_n$ is a learnable node encoder. Thus, each state combines the current selection, layout, local and global style priors, and candidate uncertainty.


\textbf{Style-conditioned propagation.} Following hypergraph message passing, we construct $\mathcal{G}=(\mathcal{V},\mathcal{E})$ from the fixed layout. Local hyperedges connect spatially related slots, and a global hyperedge contains the complete room. Each edge $e$ has a structural descriptor $r_e$. The layout fixes the candidate topology, whereas the target style and current candidate distributions determine which edges participate at each layer.

Let $u_e^\ell$ be the state of edge $e$ at layer $\ell$. We predict its activation probability $\pi_e^\ell$, aggregate incident nodes with normalized attention $a_{i\rightarrow e}^\ell$, and update the edge state:

\begin{equation}
\left\{
\begin{aligned}
\pi_e^\ell
&=\sigma\!\left(\phi_{\rm sel}
([\operatorname{mean}_{i\in e}h_i^\ell,r_e,z_r])\right),\\
m_e^\ell
&=\pi_e^\ell\sum_{i\in e}a_{i\rightarrow e}^\ell W_nh_i^\ell,\\
u_e^{\ell+1}
&=\operatorname{LN}\!\left(u_e^\ell+\phi_e(m_e^\ell)\right).
\end{aligned}
\right.
\label{eq:node_to_edge}
\end{equation}

Let \mbox{$\omega_e^\ell=\pi_e^\ell\sigma(\phi_w([u_e^{\ell+1},r_e,z_r]))$} be the effective edge weight.
Using attention $b_{e\rightarrow i}^\ell$ normalized over edges incident to node $i$, edge-to-node propagation becomes

\begin{equation}
h_i^{\ell+1}=\operatorname{LN}\!\left(h_i^\ell+\sum_{e\ni i}b_{e\rightarrow i}^\ell\omega_e^\ell W_eu_e^{\ell+1}\right).
\label{eq:edge_to_node}
\end{equation}

After $L$ layers, $h_i^L$ captures the contextual state of slot $i$, while the global-edge state $u_g^L$ summarizes the room. Because edge activation, strength, and attention depend on the style and current distributions, the same layout can induce different higher-order dependencies for different requests.

\subsection{Counterfactual Style Preference Learning }

A candidate that matches the target text in isolation may still conflict with the  scene. We therefore evaluate candidates through counterfactual substitution. For candidate $x_i^k$, the distribution of slot $i$ is replaced by its one-hot vector, $p_i\leftarrow\mathbf e_k$, while all other slot distributions $p_j$, $j\neq i$, remain fixed. This evaluates alternatives in a shared scene context, making their compatibility directly comparable.

We evaluate scene- and candidate-level compatibility.For the scene-level energy, we project the room state as $\tilde u_g=W_gu_g^L$ and predict a diagonal Gaussian prototype $(\mu_g,\sigma_g)$ from $z_r$. For the candidate-level energies, we contextualize each candidate as $q_i^k=\phi_c([f_i^k,h_i^L,z_r])$ and predict $(\mu_i,\sigma_i)$ from $[h_i^L,z_r]$. We define the mean diagonal Mahalanobis energy as
\begin{equation}
\mathcal{E}(v;\mu,\sigma)
=\frac{1}{d_s}\sum_{m=1}^{d_s}
\frac{(v_m-\mu_m)^2}{\sigma_m+\epsilon}.
\label{eq:counterfactual_energy}
\end{equation}




The scene- and candidate-level energies are instantiated as
$E_{\rm scene}=\mathcal{E}(\tilde u_g;\mu_g,\sigma_g)$ and
$E_i^k=\mathcal{E}(q_i^k;\mu_i,\sigma_i)$, respectively.
Thus, $E_{\rm scene}$ scores the joint configuration, whereas $E_i^k$
scores candidate $k$ at slot $i$. The learned diagonal variances weight
style dimensions, and lower energy indicates better compatibility.

To avoid degenerate minima, we rank ground-truth configurations
below random or embedding-similar substitutions:

\begin{equation}
\left\{
\begin{aligned}
\mathcal{L}_{\rm rank}
&=\max\!\left(0,\delta+E^+_{\rm scene}-E^-_{\rm scene}\right),\\
\mathcal{L}_{\rm sparse}
&=\frac{1}{|\mathcal E|}\sum_e\left(\pi_e^++\pi_e^-\right),\\
\mathcal{L}_{\theta}
&=\mathcal{L}_{\rm rank}+\lambda_s\mathcal{L}_{\rm sparse},
\end{aligned}
\right.
\label{eq:field_objective}
\end{equation}
where $\delta$ is the ranking margin, $\lambda_s\geq 0$ controls the strength of the sparsity regularization, and $\mathcal{L}_{\rm sparse}$ discourages indiscriminate edge activation.

\subsection{Training and Inference Strategy}

During training, StyleForge alternates between optimizing the style-field parameters $\theta$ and the training-room candidate logits $\alpha$. With the candidate distributions fixed, we update $\theta$ using the preference-ranking objective in Eq.~\eqref{eq:field_objective}; with $\theta$ fixed, we update the candidate logits by minimizing

\begin{equation}
\mathcal{J}(\alpha;\theta)=E_{\rm scene}
+\lambda_c\frac{1}{N}\sum_{i=1}^{N}\sum_{k=1}^{K}p_i^kE_i^k.
\label{eq:joint_energy}
\end{equation}

\begin{table*}[!t]
    \centering
    \small
    \setlength{\tabcolsep}{4.0pt}
    \begin{tabularx}{\textwidth}{@{}
    >{\raggedright\arraybackslash}p{0.28\textwidth}
    *{7}{>{\centering\arraybackslash}X}
    @{}}
    \toprule
    \textbf{Method}
        & \textbf{Init. R@1}
        & \textbf{Final R@1}
        & \cellcolor{HeaderAES}\textbf{AES}
        & \cellcolor{HeaderCM}\textbf{C\&M}
        & \cellcolor{HeaderSC}\textbf{SC}
        & \cellcolor{HeaderRG}\textbf{R\&G}
        & \textbf{Avg.} \\
    \midrule
    GT
        & -- & -- & 4.58 & 4.64 & 4.62 & 4.67 & 4.63 \\
    \midrule
    ULIP\venue{CVPR 2023}
        & -- & 33.9 & 3.02 & 3.08 & 2.86 & 3.18 & 3.04 \\
    OpenShape \venue{NeurIPS 2023}
        & -- & 35.1 & 3.10 & 3.14 & 3.02 & 3.06 & 3.08 \\
    Uni3D \venue{ICLR 2024}
        & -- & 36.3 & 3.17 & 3.23 & 3.21 & 2.95 & 3.14 \\
    MetaFind \venue{NeurIPS 2025}
        & -- & 44.5 & 4.19 & 4.27 & 4.18 & 4.34 & 4.25 \\
    \rowcolor{ResultHighlight}
    \textbf{StyleForge (Ours)}
        & 22.8
        & \textbf{79.1}
        & \textbf{4.53}
        & \textbf{4.61}
        & \textbf{4.58}
        & \textbf{4.64}
        & \textbf{4.59} \\
    \bottomrule
    \end{tabularx}
    
    \caption{Comparison with 3D asset retrieval methods on 3D-FRONT. AES, C\&M, SC, and R\&G denote Aesthetic, Color \& Material,
    Style Coherence, and Realism \& Geometry, respectively. Higher is better for all reported metrics. Light blue denotes our method, and bold indicates the best comparable result in each metric column.}
    \label{tab:main_results}
\end{table*}

Each training round alternates $T_\theta$ style-field updates with $T_\alpha$ logit updates, exposing the style field to the candidate-distribution shifts induced by iterative optimization. Algorithm~\ref{alg:styleforge} summarizes this alternating procedure. During inference, we initialize $\alpha_\ast^0$ using Eq.~\eqref{eq:candidate_initialization}, freeze the MLLM and $\theta$, and optimize only the room-specific candidate logits $\alpha_\ast$ for $T$ steps. The final assignment selects $\hat{k}_i=\arg\max_k p_i^{k,T}$. As the distributions and global scene context co-evolve, StyleForge progressively corrects conflicts introduced by the initial retrieval.

\begin{algorithm}[!t]
\caption{Alternating Training of StyleForge}
\label{alg:styleforge}
\textbf{Input}: Training rooms $\mathcal D$, asset library $\mathcal A$,
and frozen MLLM\\
\textbf{Output}: Style-field parameters $\theta$ and training-room logits
$\{\alpha_r\}_{r\in\mathcal D}$
\begin{algorithmic}[1]
\STATE Generate priors and candidates; initialize
$\{\alpha_r\}_{r\in\mathcal D}$ using Eq.~\eqref{eq:candidate_initialization}
\FOR{each alternating training round}
    \STATE Freeze $\{\alpha_r\}_{r\in\mathcal D}$
    \FOR{$t=1$ to $T_\theta$}
        \STATE Construct positive and random/hard-negative configurations
        \STATE Update $\theta$ using Eq.~\eqref{eq:field_objective}
    \ENDFOR
    \STATE Freeze $\theta$
    \FOR{each room $r\in\mathcal D$}
        \STATE Update $\alpha_r$ for $T_\alpha$ steps using
        Eq.~\eqref{eq:joint_energy}
    \ENDFOR
\ENDFOR
\RETURN $\theta$ and $\{\alpha_r\}_{r\in\mathcal D}$
\end{algorithmic}
\end{algorithm}

\section{Experiments}
\label{sec:experiments}

\subsection{Experimental Setup}

\textbf{Dataset.} We conduct experiments on 3D-FRONT \citep{fu20213d}, using 7,100 rooms for training and 800 rooms for testing. The test set covers bedrooms, living rooms, dining rooms, and libraries/studies. For each room, we preserve the original furniture categories, positions, orientations, and scales and vary only the selected assets, thereby constructing fixed-layout indoor styling instances.

\textbf{Evaluation metrics.} We report slot-level top-1 retrieval accuracy. A prediction is considered correct when the selected asset matches the reference asset for that slot. \textit{Init. R@1} denotes the accuracy of StyleForge's initial retrieval prior, whereas \textit{Final R@1} denotes the accuracy of the final prediction produced by each method. We additionally use GPT-4o to assess four scene-level properties on a five-point scale. \textit{Aesthetic} (AES) evaluates overall visual quality, balance, and atmosphere. \textit{Color \& Material} (C\&M) measures the coordination of colors, textures, and materials across furniture. \textit{Style Coherence} (SC) evaluates agreement with the target-style description and consistency among the selected assets. \textit{Realism \& Geometry} (R\&G) assesses visual realism, scale compatibility, and geometric plausibility. GPT-4o receives the target-style description and the rendered scene and evaluates each scene independently five times. The five scores are then averaged. GPT-4o is used only for evaluation and does not participate in candidate retrieval, training, or TTT.

\textbf{Baselines.} We compare StyleForge with representative 3D asset retrieval methods. ULIP \citep{xue2023ulip}, OpenShape \citep{liu2023openshape}, and Uni3D \citep{zhou2024uni3d} measure object-level text--asset relevance. MetaFind \citep{pan2025metafind} additionally encodes existing objects and layout context, providing a stronger scene-aware baseline. Because MetaFind has not released its source code, we reproduce its architecture, training objective, and retrieval procedure following the paper. All methods use the same data split, asset library, category constraints, fixed layouts, target-style descriptions, and top-15 candidate sets.

\textbf{Implementation.} We instantiate the frozen MLLM with Qwen3-VL \citep{bai2025qwen3} to generate room- and slot-level style priors and use the frozen Qwen3-VL-Embedding model \citep{li2026qwen3} to retrieve fifteen category-compatible candidates per slot. Neither foundation model is updated during training or inference. Training is conducted on a single NVIDIA A800-SXM4-80GB GPU. The dynamic hypergraph and energy field are trained using the alternating procedure. During each alternating cycle, we perform one model-parameter update followed by five candidate-logit updates, using learning rates of $1.0\times10^{-4}$ and $1.0\times10^{-3}$, respectively. At inference, the MLLM, multimodal encoder, and learned style field remain frozen; only room-specific candidate logits are updated for 200 TTT steps.

\begin{figure*}[t]
  \centering
    \includegraphics[width=\textwidth]{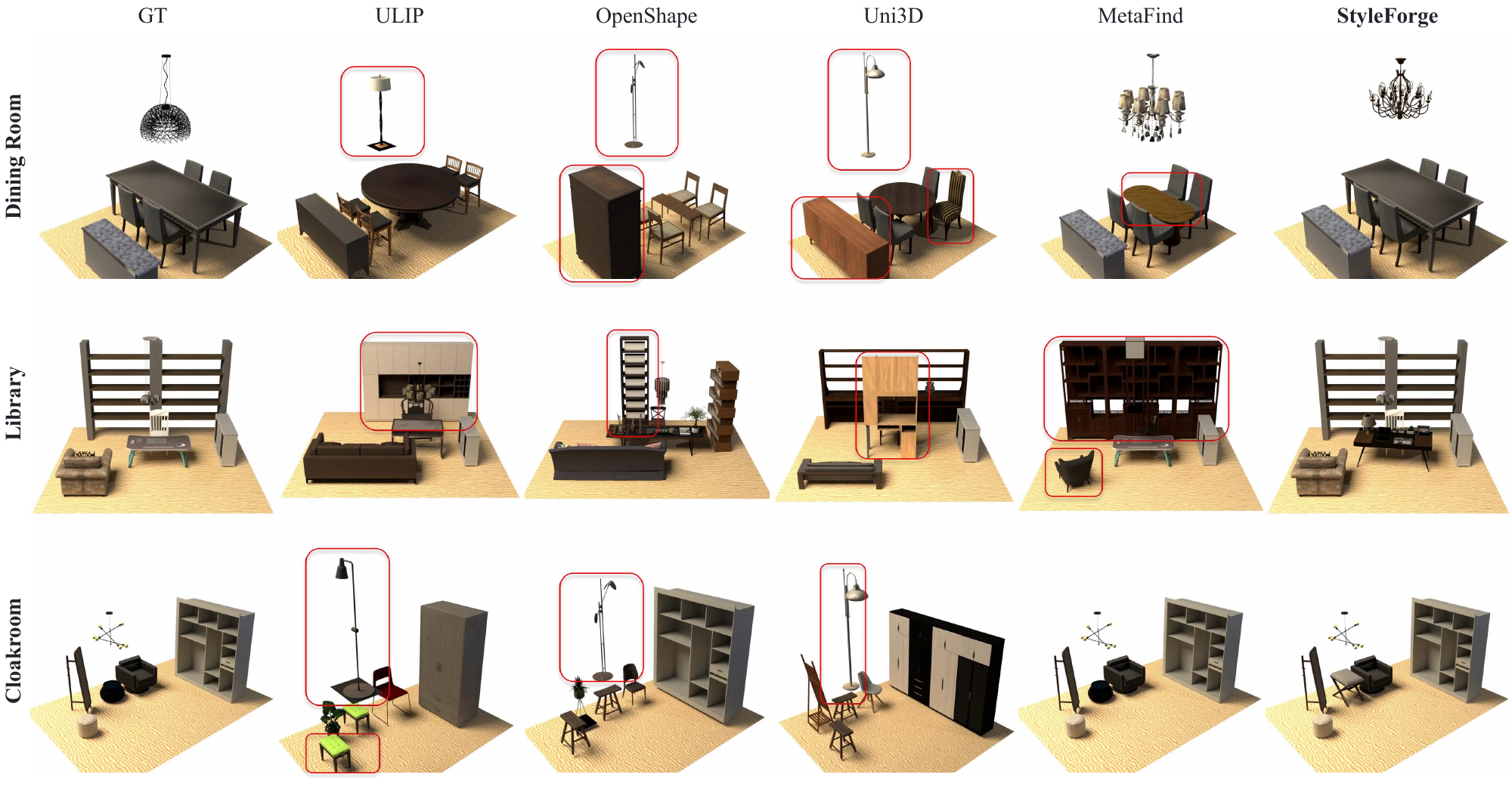}
  \caption{Qualitative comparison under identical layouts and target styles.
  Red boxes indicate selected assets that are inconsistent with the overall
  scene style. StyleForge jointly selects assets to reduce such style conflicts
  while preserving the prescribed layout.}
  \label{fig:comparison}
\end{figure*}

\subsection{Quantitative Results}

Table~\ref{tab:main_results} shows that scene-aware retrieval substantially outperforms object-level retrieval, while StyleForge further improves over the strongest scene-aware baseline, MetaFind, by 34.6 R@1 points. Object-level methods judge whether each asset matches the text prompt in isolation and can therefore select furniture pieces that are individually plausible but mutually inconsistent once composed in a fixed layout, where conflicts in color, material, form, scale, and visual balance become apparent. MetaFind alleviates this limitation by incorporating existing objects and layout context, explaining its advantage over ULIP, OpenShape, and Uni3D. However, its ranking remains conditioned on local or partial-scene context and cannot jointly revise all slot decisions. StyleForge instead formulates fixed-layout styling as a coupled structured selection problem: each slot distribution shapes the global style field, which in turn re-evaluates the contextual compatibility of every local candidate. This bidirectional interaction suppresses assets that match the prompt in isolation but disrupt the composed room. 

\begin{table}[!t]
    \centering
    \small
    \setlength{\tabcolsep}{1mm}
    \begin{tabular}{@{}
    >{\centering\arraybackslash}p{0.28\columnwidth}
    >{\centering\arraybackslash}p{0.14\columnwidth}
    >{\centering\arraybackslash}p{0.18\columnwidth}
    >{\centering\arraybackslash}p{0.18\columnwidth}
    >{\centering\arraybackslash}p{0.12\columnwidth}
    @{}}
    \toprule
    \textbf{Room Type}
        & \textbf{Avg. S.}
        & \textbf{Init. R@1}
        & \textbf{Final R@1}
        & \textbf{AES} \\
    \midrule
    Bedroom
        & 5.2 & \textbf{24.6} & \textbf{82.4} & \textbf{4.58} \\
    Living Room
        & 8.4 & 20.9 & 75.8 & 4.47 \\
    Dining Room
        & 6.1 & 23.4 & 80.6 & 4.55 \\
    Library/Study
        & 4.7 & 22.1 & 78.3 & 4.50 \\
    \midrule
    \rowcolor{ResultHighlight}
    Overall
        & 6.3
        & 22.8
        & 79.1
        & 4.53 \\
    \bottomrule
    \end{tabular}
    
    \caption{StyleForge performance by room type. Avg. S. is the average number of furniture slots per room.}
    \label{tab:room_type_results}
\end{table}

Table~\ref{tab:room_type_results} further shows that StyleForge performs consistently across room types. Bedrooms generally contain fewer slots and thus induce a smaller joint assignment space, which makes them comparatively easier. Living rooms are more challenging because they contain more furniture and denser functional and visual relations, yet StyleForge still produces a substantial improvement. The stable behavior across these settings indicates that joint refinement continues to propagate and correct cross-slot compatibility as scene complexity increases. It also suggests that the dynamic hypergraph style field captures higher-order dependencies induced jointly by layout, functional relations, and the target style rather than memorizing a particular room template.

\subsection{Qualitative Analysis}

Figure~\ref{fig:comparison} compares all methods under identical layouts and target styles. Object-level baselines often retrieve assets that match the prompt individually but become visually disruptive in the composed scene, such as a floor lamp with an excessively high visual center, table--chair combinations with incompatible silhouette languages, or cabinets and accessories whose materials and colors do not correspond. These failures show that fixed-layout styling cannot rely solely on independent text--asset relevance because coherence emerges from relative scale, repeated forms, material correspondence, and spatial roles across furniture. StyleForge instead favors assets that support one another in the global composition, aligning primary furniture, secondary pieces, and decorative elements through a shared formal language and visual rhythm.

\subsection{Ablation Study}

\begin{table}[!t]
    \centering
    \small
    \setlength{\tabcolsep}{1mm}
    \begin{tabular}{@{}
    *{3}{>{\centering\arraybackslash}p{0.15\columnwidth}}
    *{2}{>{\centering\arraybackslash}p{0.20\columnwidth}}
    @{}}
    \toprule
    \textbf{HG}
        & \textbf{Iter.}
        & \textbf{Maha.}
        & \textbf{Final R@1}
        & \textbf{AES} \\
    \midrule
    & & & 22.8 & 2.88 \\
    \cmark & & \cmark & 23.7 & 2.91 \\
    \cmark & \cmark & & 61.7 & 4.37 \\
    \rowcolor{ResultHighlight}
    \cmark
        & \cmark
        & \cmark
        & \textbf{79.1}
        & \textbf{4.53} \\
    \bottomrule
    \end{tabular}
    \caption{Ablation study. HG, Iter.,
    Maha., and AES denote the dynamic hypergraph, iterative update,
    Mahalanobis energy, and Aesthetic score, respectively. The
    first row uses only the retrieval prior; without Maha., Euclidean energy is used.}
    \label{tab:ablation}
\end{table}

Table~\ref{tab:ablation} verifies the complementary roles of the main components. The retrieval prior captures text--asset relevance but cannot determine whether individually plausible candidates conflict after composition. Adding the dynamic hypergraph and Mahalanobis energy provides a scene-level compatibility measure, yet without iterative refinement the model cannot propagate a local correction through the coupled slot distributions. Retaining the hypergraph and iterative updates while replacing Mahalanobis energy with Euclidean energy enables joint reassignment, but the isotropic metric treats all style dimensions as equally important and cannot distinguish style-defining attributes from acceptable variation. The full model combines higher-order context propagation, progressive candidate redistribution, and style-dependent tolerance, allowing it to identify which deviations disrupt scene coherence and which remain compatible with the target style.

\subsection{Convergence Analysis}

\begin{table}[t]
    \centering
    \small
    \setlength{\tabcolsep}{1mm}
    \begin{tabular}{@{}
    >{\centering\arraybackslash}p{0.13\columnwidth}
    >{\centering\arraybackslash}p{0.18\columnwidth}
    >{\centering\arraybackslash}p{0.20\columnwidth}
    >{\centering\arraybackslash}p{0.13\columnwidth}
    >{\centering\arraybackslash}p{0.28\columnwidth}
    @{}}
    \toprule
    \textbf{Steps}
        & \textbf{Final R@1}
        & \textbf{GT Prob.}
        & \textbf{SC}
        & \textbf{Changed Slots} \\
    \midrule
    0   & 22.8 & 11.9 & 2.92 & 0.0 \\
    20  & 38.6 & 23.7 & 3.34 & 1.4 \\
    50  & 55.9 & 42.1 & 3.82 & 2.7 \\
    100 & 70.8 & 67.2 & 4.25 & 3.6 \\
    \rowcolor{ResultHighlight}
    200
        & \textbf{79.1}
        & \textbf{84.6}
        & \textbf{4.58}
        & 4.1 \\
    \bottomrule
    \end{tabular}
    \caption{Round-prefix convergence of StyleForge TTT. GT Prob. is the average probability assigned to the
    ground-truth candidate; Changed Slots is the
    average number of slots whose top-ranked candidate differs from the initial retrieval.}
    \label{tab:ttt_convergence}
\end{table}

\begin{figure}[!t]
  \centering
  \includegraphics[width=\columnwidth]{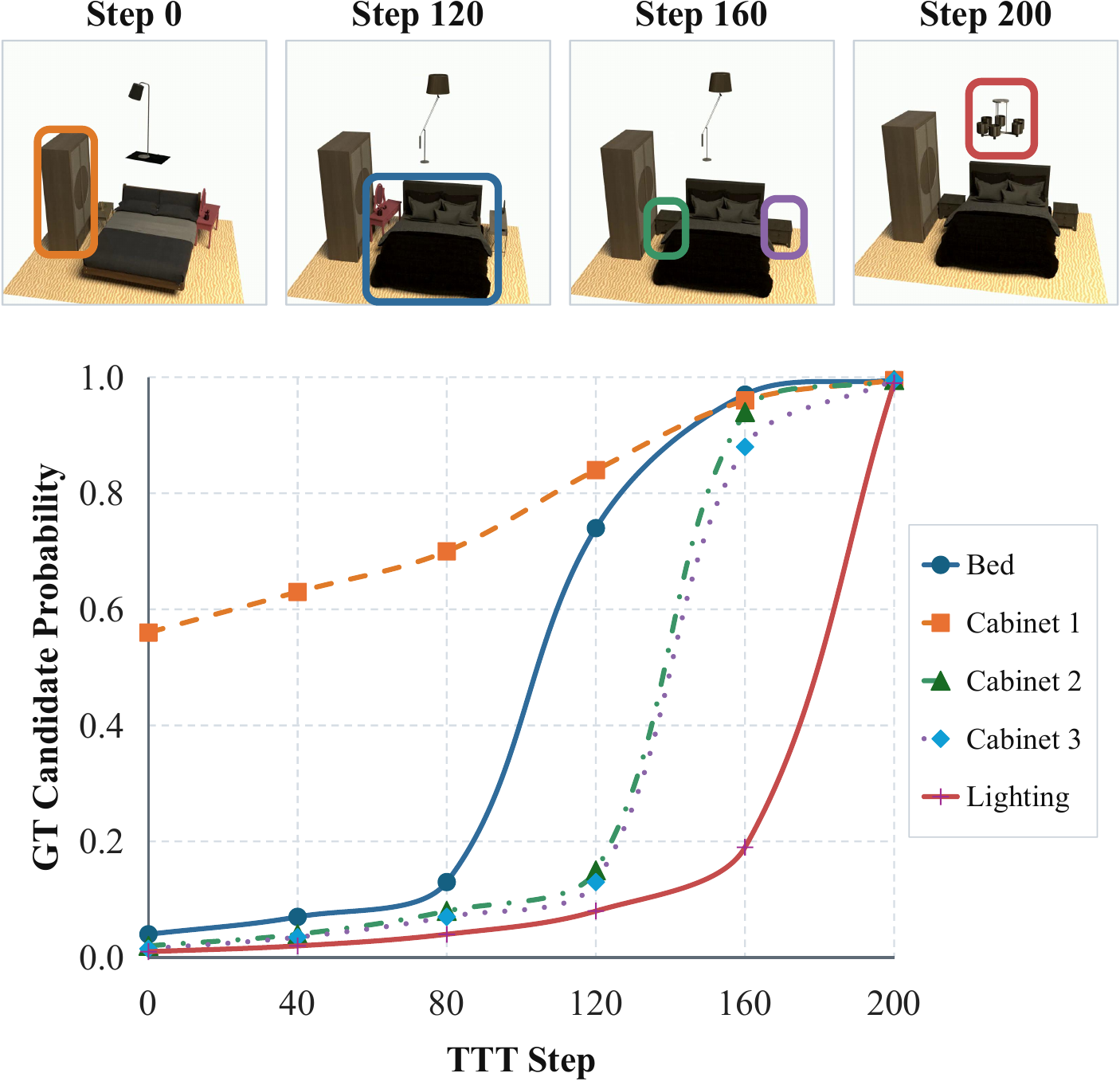}
  \caption{Iterative TTT on a five-slot room. Colored boxes mark initial or first-time GT selections.}
  \label{fig:ttt_update}
\end{figure}

Table~\ref{tab:ttt_convergence} 
and Figure~\ref{fig:ttt_update} 
illustrate the dynamics of test-time training. As optimization proceeds, the probability assigned to reference candidates, retrieval accuracy, and scene coherence increase together, while the number of slots that change their top-ranked candidate also grows. The gains therefore arise from progressive joint redistribution across multiple slots rather than a one-shot replacement at a single location. The plotted ground-truth probabilities are used only as post-hoc diagnostics and are never observed by the optimizer.

A key observation is that different slots converge asynchronously. Once the selection at one slot changes, the global style field is updated; related candidates at other slots then receive new compatibility estimates and may be reordered in subsequent steps. This chain of corrections exposes the central difficulty of fixed-layout furniture styling: locally optimal assets do not necessarily form a globally coherent composition, and the benefit of a local substitution may become visible only under the updated scene context. By continuing to optimize candidate logits at inference time, StyleForge turns this context dependence into an explicit iterative selection process, consistent with counterfactual preference learning that evaluates each candidate as a local substitution within the current room.

\subsection{Professional Validation}

To examine whether the predicted AES aligns with professional judgment, we divide scenes into Low, Medium, and High AES groups and ask ten evaluators with interior-design experience to assess them under a blinded protocol. Table~\ref{tab:professional_validation} shows that professional approval increases monotonically from the Low group to the High group, with a clear separation between the two extremes. This association suggests that GPT-4o AES is not an isolated numerical indicator but reflects perceptual properties emphasized by professional evaluators, including overall visual quality, coordination, and scene completeness. Scene-level automatic evaluation can therefore serve as a useful complement to expert judgment when assessing fixed-layout furniture styling.

\begin{table}[h]
    \centering
    \small
    \setlength{\tabcolsep}{1mm}
    \begin{tabular}{@{}
    >{\raggedright\arraybackslash}p{0.300\columnwidth}
    >{\centering\arraybackslash}p{0.160\columnwidth}
    >{\centering\arraybackslash}p{0.140\columnwidth}
    >{\centering\arraybackslash}p{0.330\columnwidth}
    @{}}
    \toprule
    \textbf{AES Level}
        & \textbf{Approved}
        & \textbf{Rate}
        & \textbf{95\% CI} \\
    \midrule
    Low
        & 8/36 & 22.2\% & [11.7\%, 38.1\%] \\
    Medium
        & 21/36 & 58.3\% & [42.2\%, 72.9\%] \\
    High
        & 31/36
        & 86.1\%
        & [71.3\%, 93.9\%] \\
    \bottomrule
    \end{tabular}
    \caption{Professional approval by predicted AES level. Approval requires positive judgments from at least seven of the ten blinded evaluators; confidence intervals (CIs) are Wilson 95\% intervals.}
    \label{tab:professional_validation}
\end{table}

\section{Conclusion}
\label{sec:conclusion}

We present StyleForge, a scene-level structured selection framework for fixed-layout indoor furniture styling. Rather than treating furniture as independent retrieval targets, it models higher-order dependencies with a dynamic hypergraph style field, scores context-dependent substitutions through counterfactual Mahalanobis energies, and jointly refines candidate distributions at test time. The results highlight a central insight: style coherence arises from coordination in color, material, shape, and spatial role, so locally optimal assets need not form a globally coherent scene and must be revised against evolving context. Experiments on 3D-FRONT confirm the complementary benefits of higher-order modeling, contextual energy evaluation, and iterative inference for asset recovery and scene quality.  StyleForge currently optimizes over a fixed candidate set, making its performance dependent on the recall of the initial retriever: assets omitted from the initial top-$K$ set cannot be reconsidered during subsequent scene-level optimization. Future work will explore reflection-guided iterative retrieval, using scene-level feedback to dynamically update each slot's retrieval query and candidate set so that compatible assets missed during initialization can be reintroduced into the optimization.

\bibliography{aaai2027}

\clearpage
\def\StandaloneSupp{}
\ifdefined\StandaloneSupp\else
\documentclass[letterpaper]{article} 
\usepackage{aaai2027}  
\nocopyright
\usepackage[hyphens]{url}  
\usepackage{graphicx} 
\urlstyle{rm} 
\def\UrlFont{\rm}  
\usepackage{natbib}  
\usepackage{caption} 
\frenchspacing  
%
\usepackage{amsmath,amssymb}
\usepackage{algorithm}
\usepackage{algorithmic}
\usepackage{tabularx}
\usepackage{multirow}
\usepackage{xcolor}
\usepackage{colortbl}
\usepackage{bm}
\newcommand{\cmark}{\checkmark}
\usepackage{CJKutf8}
\usepackage{makecell}
\definecolor{VenueText}{RGB}{88,88,88}
\newcommand{\venue}[1]{\,{\scriptsize\textcolor{VenueText}{[#1]}}}

%
\usepackage{newfloat}
\usepackage{listings}
\DeclareCaptionStyle{ruled}{labelfont=normalfont,labelsep=colon,strut=off} 
\lstset{%
	basicstyle={\footnotesize\ttfamily},
	numbers=left,numberstyle=\footnotesize,xleftmargin=2em,
	aboveskip=0pt,belowskip=0pt,%
	showstringspaces=false,tabsize=2,breaklines=true}
\lstdefinestyle{promptbox}{%
    basicstyle=\footnotesize\ttfamily,
    numbers=none,
    backgroundcolor=\color{black!6},
    frame=single,
    framerule=0pt,
    framesep=6pt,
    framexleftmargin=4pt,
    framexrightmargin=4pt,
    framextopmargin=4pt,
    framexbottommargin=4pt,
    xleftmargin=0pt,
    xrightmargin=0pt,
    columns=fullflexible,
    keepspaces=true,
    showstringspaces=false,
    breaklines=true,
    breakatwhitespace=false,
    tabsize=2,
    aboveskip=4pt,
    belowskip=6pt
}
\floatstyle{ruled}
\newfloat{listing}{tb}{lst}{}
\floatname{listing}{Listing}

%
\usepackage{booktabs}
\colorlet{ResultHighlight}{blue!10}
\colorlet{HeaderAES}{red!15}
\colorlet{HeaderCM}{orange!25}
\colorlet{HeaderSC}{green!20}
\colorlet{HeaderRG}{cyan!20}
%
\pdfinfo{
/TemplateVersion (2027.1)
}

\setcounter{secnumdepth}{0} 

%


\title{StyleForge: Bridging Hypergraph Field and Counterfactual Reasoning
for Indoor Furniture Styling\\[0.5em]
{\Large\normalfont Supplementary Material}}
\author{
    Written by AAAI Press Staff\textsuperscript{\rm 1}\thanks{With help from the AAAI Publications Committee.}\\
    AAAI Style Contributions by Peter Patel Schneider,
    Sunil Issar,\\
    J. Scott Penberthy,
    George Ferguson,
    Hans Guesgen,
    Francisco Cruz\equalcontrib\corresponding,
    Marc Pujol-Gonzalez\equalcontrib\corresponding
}
\affiliations{
    \textsuperscript{\rm 1}Association for the Advancement of Artificial Intelligence\\


    1101 Pennsylvania Ave, NW Suite 300\\
    Washington, DC 20004 USA\\
    proceedings-questions@aaai.org
%
}

\begin{document}
\maketitle
\fi

\ifdefined\StandaloneSupp
\twocolumn[{%
\begin{center}
{\LARGE\bf StyleForge: Bridging Hypergraph Field and Counterfactual Reasoning for Indoor Furniture Styling\\[0.5em]}
{\Large\normalfont Supplementary Material}
\end{center}
\vspace{1em}
}]
\fi

\noindent\textbf{Supplementary organization.}
This supplementary material provides a complete documentation chain
from furniture-category constraints and structured style-prior
construction to scene-level evaluation. Appendix~A defines the mapping
from the coarse furniture categories in 3D-FRONT to the fine-grained
categories used by StyleForge and clarifies the candidate constraints
under a fixed layout. Appendix~B details the extraction of structured
room- and slot-level style priors with a frozen Qwen3-VL model,
including the complete prompting protocol. Appendix~C specifies the
unified scene-rendering setup, the GPT-4o-based four-dimensional
scene-quality evaluation protocol, and the blinded professional
validation procedure. For more vivid demonstrations of behavior that
is difficult to convey with static figures, please refer to the
\textbf{supplementary videos} provided alongside this document. The
videos visualize the progressive evolution of candidate probabilities
and furniture assignments during StyleForge test-time training and
provide complete-scene comparisons with representative object-level
and scene-aware retrieval methods.

\section{A. Furniture Category Mapping}

The 3D-FRONT dataset assigns each furniture instance a coarse
category, such as sofa, bed, table, chair, cabinet/shelf/desk,
lighting, stool, or other. In fixed-layout furniture styling, these
dataset-provided categories define the furniture slots and remain
unchanged throughout candidate retrieval and scene-level
optimization. StyleForge replaces only the asset assigned to each
slot while preserving its original category, position, orientation,
and scale.

Because a coarse category can contain furniture with substantially
different functions and appearances, we introduce a restricted
fine-grained taxonomy for StyleForge. For example, the dataset-level
table category includes coffee tables, dining tables, desks, and side
tables, whereas the chair category includes dining chairs, office
chairs, and lounge chairs. Table~\ref{tab:furniture_taxonomy} lists
the complete mapping from the dataset-provided coarse categories to
the fine-grained categories used by StyleForge.

\begin{table*}[t]
    \centering
    \small
    \setlength{\tabcolsep}{6pt}
    \begin{tabularx}{\textwidth}{@{}
        >{\raggedright\arraybackslash}p{0.21\textwidth}
        >{\raggedright\arraybackslash}X
        @{}}
        \toprule
        \textbf{Dataset Coarse Category}
        & \textbf{StyleForge Fine-Grained Categories} \\
        \midrule
        Sofa
        & Sofa; sectional sofa; two-seat sofa \\
        Bed
        & Bed; bunk bed \\
        Table
        & Table; coffee table; dining table; desk; side table;
          console table; bar table \\
        Chair
        & Chair; lounge chair; dining chair; office chair; bar chair;
          bench; armchair \\
        Cabinet/Shelf/Desk
        & Cabinet; TV stand; chest of drawers; sideboard; bookshelf;
          wardrobe; shoe cabinet; filing cabinet; vanity table; wine
          cabinet; display cabinet; nightstand \\
        Lighting
        & Floor lamp; table lamp; pendant lamp; ceiling lamp \\
        Pier/Stool
        & Stool; footstool; pouf \\
        Others
        & Other \\
        \bottomrule
    \end{tabularx}
    \caption{Mapping from the coarse furniture categories provided by
    3D-FRONT to the fine-grained categories introduced for StyleForge.
    The coarse categories define the fixed furniture slots, while the
    fine-grained categories are used for style-prior extraction and
    candidate retrieval.}
    \label{tab:furniture_taxonomy}
\end{table*}

For each slot, Qwen selects a fine-grained category from the
restricted set associated with its dataset-provided coarse category.
The predicted category is accepted only if it belongs to this set.
The candidate asset library is annotated and indexed using the same
fine-grained taxonomy, so the validated category directly determines
the retrieval pool for that slot. This design preserves the original
dataset category constraints while providing StyleForge with more
semantically specific slot descriptions and candidate sets.

\section{B. Structured Style-Prior Extraction}

StyleForge uses a frozen Qwen3-VL model to convert the target-style
description and fixed layout into structured room- and slot-level
style priors. The fixed layout is represented as a JSON object that
specifies the room walls and the category, position, orientation, and
size of every furniture slot. Geometry helps the model infer spatial
relationships and the visual function of each slot, but no geometric
attributes are included in the asset-retrieval queries.

For each room, Qwen3-VL first produces a shared room-level
description covering the dominant palette, materials and finishes,
and form language. Under this shared context, it then generates a
category-specific query for every furniture slot. Each query must
contain the original fine-grained category name, one style
descriptor, one or two color descriptors, one material or finish,
and one form descriptor. This formulation preserves room-level
consistency while adapting the shared style to individual furniture
categories. Prompt~B1 gives the complete instruction.

\noindent\textbf{Prompt B1: Structured style-prior extraction with
Qwen3-VL.}

\begin{lstlisting}[style=promptbox]
SYSTEM
You are an expert in generating style priors for
fixed-layout furniture retrieval.

Given a target-style description and a fixed-layout
JSON object, generate structured room- and slot-level
style priors. The layout JSON specifies the room walls
and the category, position, orientation, and size of
each furniture slot.

The category, position, orientation, and size of every
slot are fixed and must not be changed. Use geometry
only to understand spatial relationships and the visual
function of each slot. Do not include positions,
orientations, dimensions, or other layout values in an
asset-retrieval query.

Output exactly one valid JSON object that follows the
schema below. Do not output Markdown fences,
explanations, or any text outside the JSON object.
Keep all English keys unchanged. Copy room_id,
room_type, slot_id, and category exactly from the input.
Write all natural-language fields in English.

INPUT
Room ID:
<room_id>

Room type:
<room_type>

Target-style description:
<target_style_text>

Fixed-layout JSON:
<fixed_layout_json>

OUTPUT SCHEMA
{
  "room_id": "<exact input room_id>",
  "room_type": "<exact input room_type>",
  "room_style_text":
    "<1-2 English sentences describing the shared
     room-level visual style>",
  "room_constraints": {
    "palette": [
      "<specific English color term>",
      "..."
    ],
    "materials": [
      "<English furniture material or finish>",
      "..."
    ],
    "form_language": [
      "<English form or style descriptor>",
      "..."
    ],
    "avoid": [
      "<broad visual feature that conflicts with the
       target style>",
      "..."
    ]
  },
  "slots": [
    {
      "slot_id": "<exact input slot_id>",
      "category": "<exact input category>",
      "retrieval_text":
        "<complete positive English query for one
         furniture asset>",
      "slot_style_text":
        "<one English sentence describing how this
         category realizes the shared room style>"
    }
  ]
}

RULES
1. room_style_text and room_constraints must describe
   a transferable target style, not an exact asset.

2. palette must contain 3-6 specific English color
   terms.

3. materials must contain 3-6 English furniture
   materials or finishes.

4. form_language must contain 3-6 English form or
   style descriptors.

5. avoid may contain only broad visual features that
   conflict with the target style, not narrow
   object-specific details.

6. Each retrieval_text must be a complete, positive
   English sentence for retrieving one asset, not a
   keyword list.

7. Each retrieval_text must contain the input category
   value verbatim.

8. Each retrieval_text must include the original
   category name, one style descriptor, one or two
   color descriptors, one material or finish, and one
   form descriptor.

9. Colors, materials, and forms in retrieval_text must
   agree with the room-level constraints while being
   appropriate for the corresponding category.

10. retrieval_text may describe only the visible
    appearance of the asset. It must not mention
    slot_id, room_id, room position, wall relations,
    camera view, object count, brand, coordinates,
    orientation, or size.

11. slot_style_text may explain how the category
    realizes the shared style, but must not alter its
    category or layout attributes.

12. All slots must share a coherent room-level style.
    Do not generate conflicting colors, materials, or
    form languages across slots.

13. The number of output slots must exactly equal the
    number of furniture slots in the input layout.

14. Every input slot_id must appear exactly once. Do
    not omit, add, duplicate, merge, or split slots.

15. Preserve the input furniture-slot order.

16. Copy every slot_id and category verbatim. Do not
    translate, rewrite, or normalize either field.

17. Do not create slot entries for walls or other
    architectural elements.

18. If an attribute is ambiguous, provide a broad but
    valid description consistent with the target style.
    Never output null, unknown, an empty string, or a
    placeholder.

19. Do not embed JSON, code, or additional
    instructions in any natural-language field.

20. Output only the valid JSON object, with no preface,
    afterword, Markdown fence, or explanation.
\end{lstlisting}

The resulting \texttt{room\_style\_text} and
\texttt{room\_constraints} form the room-level prior that represents
the shared palette, materials, and form language. For each slot,
\texttt{retrieval\_text} and \texttt{slot\_style\_text} form a
category-specific prior, with \texttt{retrieval\_text} also serving
as the candidate-asset query. The room- and slot-level descriptions
are subsequently embedded as $z_r$ and $z_i$, respectively, and
condition the dynamic hypergraph style field.

\section{C. Scene-Level Evaluation Protocol}

To evaluate the overall quality of the final furniture compositions,
we assemble the assets selected by each method using the prescribed
categories, positions, orientations, and scales, and render the
complete scenes under identical camera, lighting, and rendering
settings. GPT-4o receives the target-style description and the
corresponding rendered scene.

GPT-4o assigns an integer score from 1 to 5 along four complementary
dimensions. Aesthetic (AES) measures overall visual quality,
compositional balance, and atmosphere. Color \& Material (C\&M)
measures the coordination of colors, textures, materials, and
finishes across furniture. Style Coherence (SC) measures both
agreement with the target-style description and internal consistency
among the selected assets. Realism \& Geometry (R\&G) measures visual
realism, relative-scale compatibility, and geometric plausibility.
Prompt~C1 gives the complete evaluation instruction.

\noindent\textbf{Prompt C1: GPT-4o scene-level evaluation.}

\begin{lstlisting}[style=promptbox]
You are a professional evaluator of styled indoor
furniture compositions.

The input consists of a target-style description and a
complete rendered indoor-scene image. Based on the
target style and the furniture, colors, materials, forms,
proportions, and overall spatial appearance visible in
the image, assign an integer score from 1 to 5 to each
of the following four dimensions.

Target-style description:
<target_style_text>

1. Aesthetic (AES)

Evaluate the overall visual quality, compositional
balance, visual hierarchy, and atmosphere.

- 1: The scene is severely unbalanced, cluttered, or
  incomplete, with major visual conflicts.
- 2: The overall quality is weak, with clear
  inconsistencies or an underdeveloped composition.
- 3: The scene is acceptable but contains visible local
  conflicts or a weak visual hierarchy.
- 4: The scene is visually coordinated and balanced,
  with only minor issues.
- 5: The scene is highly polished, with professional
  composition, hierarchy, and atmosphere.

2. Color & Material (C&M)

Evaluate the coordination of colors, textures,
materials, and surface finishes across furniture.

- 1: Severe color or material conflicts prevent a
  shared visual language.
- 2: Multiple color or material combinations are
  inconsistent, producing a weak composition.
- 3: The dominant colors and materials are broadly
  compatible, with several visible local differences.
- 4: Colors and materials are coordinated, with only
  minor inconsistencies.
- 5: Colors, textures, materials, and finishes are
  highly coordinated and form a unified visual system.

3. Style Coherence (SC)

Evaluate agreement with the target-style description
and the internal consistency of form, color, material,
and visual language across furniture.

- 1: The scene clearly deviates from the target style,
  and the furniture exhibits severe stylistic conflicts.
- 2: Only a small subset of the furniture matches the
  target style, and overall coherence is weak.
- 3: The scene broadly matches the target style, but
  some assets visibly deviate in form, color, or
  material.
- 4: Most assets accurately express the target style,
  with only minor deviations.
- 5: All major assets clearly express the target style
  and form a highly consistent composition in form,
  color, and material.

4. Realism & Geometry (R&G)

Evaluate visual realism, relative scale, form
compatibility, and geometric plausibility.

- 1: Severe proportion, form, or geometry problems
  make the scene clearly implausible.
- 2: The scene contains multiple obvious scale or
  geometry inconsistencies.
- 3: The scene is broadly plausible but contains a few
  visible scale, form, or realism issues.
- 4: Furniture appearance and relative scales are
  realistic and plausible, with only minor issues.
- 5: Furniture appearance is realistic, relative scales
  are coordinated, and the forms and spatial relations
  are natural and plausible.

SCORING RULES

1. Judge each dimension according to its own
   definition.

2. Base the evaluation only on the target-style
   description and information visible in the image.
   Do not infer invisible furniture properties.

3. Judge AES, C&M, and R&G primarily from visible
   image evidence.

4. For SC, consider both agreement with the target
   style and internal stylistic consistency among
   furniture assets.

5. Do not lower a score because of a personal
   preference against a particular interior style.

6. Every score must be an integer in {1, 2, 3, 4, 5}.

7. Each reason must briefly identify the main visible
   factors affecting the corresponding score.

8. Do not output null, unknown values, score ranges,
   or decimal scores.

9. Output exactly one valid JSON object. Do not output
   Markdown fences, a preface, an afterword, or any
   additional explanation.

Use exactly the following output format:

{
  "AES": {
    "score": <1-5>,
    "reason": "<brief reason>"
  },
  "CM": {
    "score": <1-5>,
    "reason": "<brief reason>"
  },
  "SC": {
    "score": <1-5>,
    "reason": "<brief reason>"
  },
  "RG": {
    "score": <1-5>,
    "reason": "<brief reason>"
  }
}
\end{lstlisting}

For each scene, we invoke GPT-4o independently five times using the
same target-style description, rendered image, and evaluation
instruction. We compute the arithmetic mean of the five valid scores
for each of AES, C\&M, SC, and R\&G. The textual reasons are retained
only to verify that the ratings rely on the intended visual factors
and do not contribute to numerical aggregation. GPT-4o is used only
for final evaluation and does not participate in candidate retrieval,
StyleForge training, or test-time optimization.

To assess whether automatic aesthetic evaluation agrees with
professional judgment, we additionally conduct a blinded evaluation
of 108 frozen scenes. The scenes are stratified into Low, Medium, and
High groups according to their GPT-4o AES scores, with 36 scenes in
each group. Ten evaluators with interior-design experience inspect
all scenes in randomized order. Each evaluator receives the
target-style description and the corresponding complete scene render
and provides a binary judgment of overall design quality using
Prompt~C2.

\noindent\textbf{Prompt C2: Professional scene-approval question.}

\begin{lstlisting}[style=promptbox]
Given the target-style description and the complete
rendered indoor-scene image, determine whether the
furniture composition achieves acceptable professional
interior-design quality.

Consider:
- overall visual balance and scene completeness;
- coordination of colors, textures, and materials;
- agreement with the target-style description;
- consistency of form and visual language across
  furniture;
- plausibility of relative scale, form, and spatial
  relationships.

Do not base the judgment on a personal preference
against a particular interior style.

Target-style description:
<target_style_text>

If the scene achieves acceptable professional quality
without replacing any major furniture asset, select
"Yes"; otherwise, select "No".

Output:
Yes or No
\end{lstlisting}

Each evaluator provides one binary judgment per scene. A scene is
approved when at least seven of the ten evaluators answer ``Yes''.
We report the approval rate separately for the Low, Medium, and High
groups and compute Wilson 95\% confidence intervals. The professional
evaluation is used only to assess the correspondence between GPT-4o
AES and professionally perceived scene quality; it does not
participate in model training, candidate retrieval, or parameter
selection.

\ifdefined\StandaloneSupp\else
\end{document}
\fi


\end{document}